\documentclass[conference]{IEEEtran}
\IEEEoverridecommandlockouts
\usepackage{cite}
\usepackage{amsmath,amssymb,amsfonts}
\usepackage{algorithmic}
\usepackage{graphicx}
\usepackage{textcomp}
\usepackage{xcolor}
\def\BibTeX{{\rm B\kern-.05em{\sc i\kern-.025em b}\kern-.08em
    T\kern-.1667em\lower.7ex\hbox{E}\kern-.125emX}}
    
\usepackage{booktabs}
\usepackage{multirow}
\begin{document}

\title{Boundless Across Domains: A New Paradigm of Adaptive Feature and Cross-Attention for Domain Generalization in Medical Image Segmentation}

\author{\IEEEauthorblockN{Yuheng Xu}
\IEEEauthorblockA{\textit{College of Computer Science} \\
\textit{Chongqing University}\\
Chongqing University \\
zeruagogogo@gmail.com}
\and
\IEEEauthorblockN{Taiping Zhang}
\IEEEauthorblockA{\textit{College of Computer Science} \\
\textit{Chongqing University}\\
Chongqing University \\
tpzhang@cqu.edu.cn}
}

\maketitle

\begin{abstract}
Domain-invariant representation learning is a powerful method for domain generalization. Previous approaches face challenges such as high computational demands, training instability, and limited effectiveness with high-dimensional data, potentially leading to the loss of valuable features. To address these issues, we hypothesize that an ideal generalized representation should exhibit similar pattern responses within the same channel across cross-domain images. Based on this hypothesis, we use deep features from the source domain as queries, and deep features from the generated domain as keys and values. Through a cross-channel attention mechanism, the original deep features are reconstructed into robust regularization representations, forming an explicit constraint that guides the model to learn domain-invariant representations. Additionally, style augmentation is another common method. However, existing methods typically generate new styles through convex combinations of source domains, which limits the diversity of training samples by confining the generated styles to the original distribution. To overcome this limitation, we propose an Adaptive Feature Blending (AFB) method that generates out-of-distribution samples while exploring the in-distribution space, significantly expanding the domain range. Extensive experimental results demonstrate that our proposed methods achieve superior performance on two standard domain generalization benchmarks for medical image segmentation.
\end{abstract}


\section{Introduction}

In modern medicine, image segmentation is a crucial technology. Its primary task is to separate different tissues, organs, or lesion areas from the background in medical images, thereby providing essential support for clinical diagnosis, treatment planning, and patient monitoring. Thanks to the rapid advancements in deep learning technologies, significant progress has been made in the field of medical image segmentation. However, numerous challenges remain in practical applications. One of the main challenges stems from the differences in imaging protocols, equipment vendors, operators, and patient populations, which often lead to discrepancies between the test data (referred to as the target domain) and the training data (referred to as the source domain). This discrepancy, known as domain shift, significantly degrades the performance of existing medical image segmentation models. 
\begin{figure}
	\centering
	\includegraphics[width=\columnwidth]{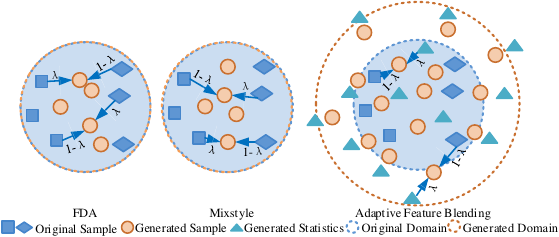}
	\caption{Visualization of synthetic feature statistics samples using MixStyle, FDA, and our proposed Adaptive Feature Blending (AFB) method.}
	\label{fig1}
\end{figure}
To address this issue, domain generalization (DG) techniques have been introduced to enhance the generalization ability of models. Existing approaches to domain generalization can be categorized into two main types. The first type focuses on learning domain-invariant features from multiple source domains \cite{liu2020shape}, while the second explicitly learns the domain shifts between multiple source domains \cite{wang2020dofe,zhou2022generalizable}. Both approaches typically rely on data augmentation to increase the diversity of the training data, thereby reducing the model's overfitting to the training data. In this context, several data augmentation methods have been proposed, such as MixStyle \cite{zhou2021domain} and FDA \cite{yang2020fda}, which generate new domain style samples by forming convex combinations of the statistical data of low-level features or the low-frequency amplitude components from different images in the source domain.
Although these methods are computationally efficient, they are limited to generating samples within the existing distribution (as shown in Fig. \ref{fig1}), which may restrict the network's generalization capability. In contrast, the research \cite{zhong2022adversarial} and \cite{zhang2023adversarial} apply adversarial attacks to feature statistics, generating challenging adversarial samples to address the aforementioned limitation, but at the cost of introducing a significant computational burden.

The study \cite{peng2021global} demonstrates that random augmentation can significantly enhance diversity and promote the learning of domain-invariant representations. However, this approach may also lead to overgeneralization or difficulty in model convergence. To address these issues, we propose a simple yet effective method called Adaptive Feature Blending(AFB), which aims to expand the domain distribution by perturbing the style information of source domain instances. We begin by randomly sampling augmentation statistics from a uniform distribution that encompasses most feature statistics. Then, we randomly mix the augmented and original statistics along the channel dimension to blend feature styles. This approach not only covers the in-distribution space but also generates out-of-distribution samples (as shown in Fig. \ref{fig1}), while introducing references from the original features, which could avoid the model from overgeneralizing or failing to converge due to excessive randomization.

Learning domain-invariant representations \cite{nguyen2021domain,niu2023knowledge} is crucial in domain generalization, as it enables models to perform well on unseen target domains by learning features that are robust across different domains. Previous domain-invariant representation learning methods face challenges such as high computational resource demands, unstable training, limited effectiveness with high-dimensional complex data, and the potential loss of useful feature information.
To overcome these challenges, we developed a display constraint based on a cross-channel attention mechanism called Dual Cross-Attention Regularization(DCAR) to learn domain-invariant representations. We hypothesize that an ideal generalized representation should exhibit similar pattern responses within the same channel across cross-domain images. Specifically, the generated images are merely style-transformed versions of the original images, with semantics unchanged. Since deep features of the encoder contain rich semantic information, the generated and original deep features should be highly similar. Based on this hypothesis, we use the deep features of the original image as queries and those of the generated image as keys and values. We then compute a similarity matrix between each channel of the original and generated features. The original image feature channels are reconstructed using the most similar channels from the generated image, transforming the original deep features into robust regularization representations. In this way, we establish a display constraint to guide the model in learning domain-invariant representations. Moreover, due to the semantic consistency between the generated images and the original images, we also use the deep features of the generated images as queries, with the original image features as keys and values, to build an additional display constraint.

Our contributions can be summarized as follows: (1)A simple yet effective data augmentation technique generates out-of-distribution samples while exploring the intra-distribution space, significantly enhancing the diversity of the training data. (2)We propose constructing a display constraint from deep features to learn domain-invariant representations. The proposed Dual Cross-Attention Regularization can be seamlessly integrated into segmentation models as a regularization module, enabling improved domain generalization performance. (3)Extensive experiments have validated the effectiveness and superiority of the proposed Adaptive Feature Blending and Dual Cross-Attention Regularization on two standard domain generalization benchmarks for medical image segmentation. 
\section{Related work}



\textbf{Domain Generalization.} In the domain generalization setting, models are required to be trained solely on source domain data and still perform well on unseen target domains \cite{chattopadhyay2020learning,chen2023treasure}. Existing methods mainly include data augmentation \cite{zhou2021domain,xu2020robust,zhang2024domain}, adversarial training \cite{zhu2022localized,sicilia2023domain}, and domain-invariant representation learning \cite{liu2020shape,liu2021feddg,li2018domain,choi2021robustnet}. Data augmentation simulates potential domain shifts by generating diverse training samples, while adversarial training enhances model robustness by creating perturbed samples. Domain-invariant representation learning aims to extract features that remain stable across different domains.


\section{Method}
\begin{figure}
	\centering
	\includegraphics[width=\columnwidth]{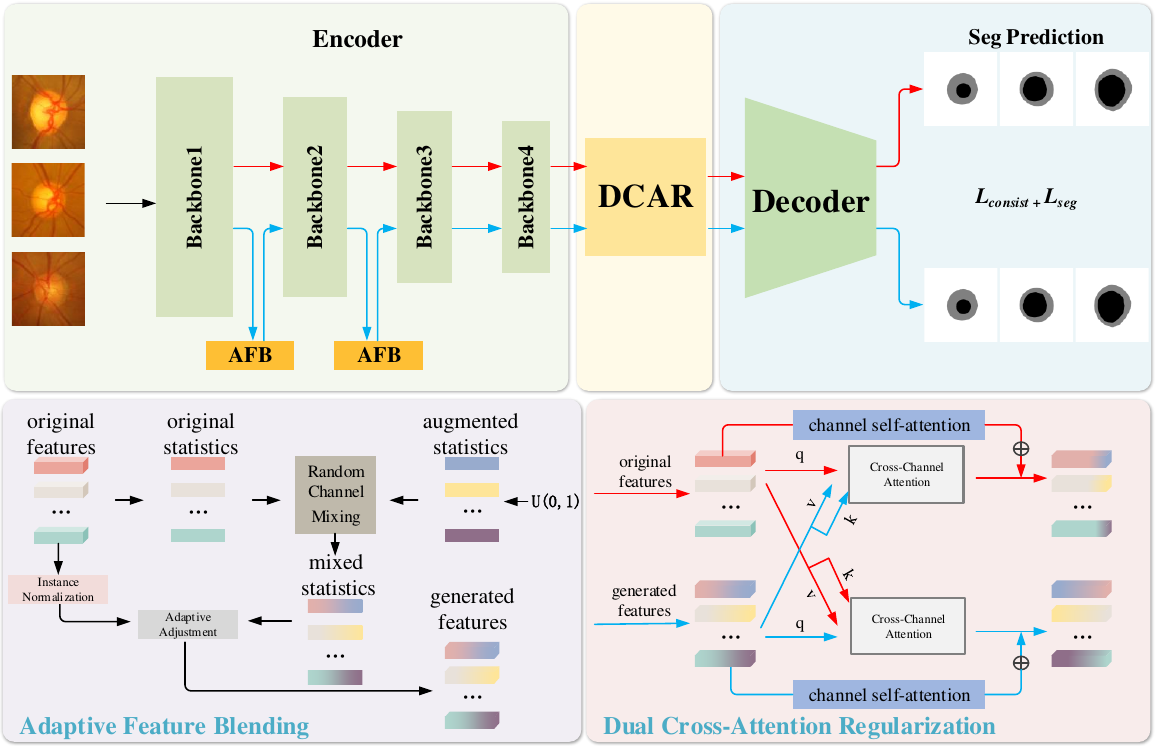}
	\caption{Overview of our proposed method. AFB generates out-of-distribution samples while exploring the in-distribution space, significantly expanding the domain range. DCAR reconstructs the original deep features and the generated deep features into robust regularization constraints through a cross-attention mechanism, thereby guiding the model to learn domain-invariant representations.}
	\label{fig2}
\end{figure}

The overview of our framework is illustrated in the Fig. \ref{fig2}. Given a set of $D$ source domains ($\left\{\left(x_{i}^{d}, y_{i}^{d}\right)_{i=1}^{N_{d}}\right\}_{d=1}^{D}$), our objective is to endow the medical image segmentation models the capability to extract domain-invariant representations. Here, $x_{i}^{d}$ is the $i^{t h}$ image from $d^{t h}$ source domain, $y_{i}^{d}$ is the segmentation label of $x_{i}^{d}$, and $N_{d}$ is the number of samples in $d^{t h}$ source domain.
\subsection{Adaptive Feature Blending}
Recent studies \cite{ulyanov2016instance,huang2017arbitrary,dumoulin2016learned} have shown that statistics of CNN feature, particularly the channel mean and standard deviation of feature maps, effectively capture the characteristics of stylized images and can be regarded as representations of visual domains. Inspired by this, we propose to regularize CNN training by perturbing the style information of source domain instances. Specifically, to cover the in-distribution space and generate out-of-distribution samples, we randomly sample augmentation statistics $ \mu^{\prime},\sigma^{\prime} \in \mathbb{R}^{B \times C}$($B$ and $C$ respectively denote the batch size and channel) from a uniform distribution that encompasses most feature statistics: $\sigma^{\prime} \sim U(0,1), \mu^{\prime} \sim U(0,1)$. We then sample $b \in \mathbb{R}^{B \times C}$ from a $Beta$ distribution: $b \sim \operatorname{Beta}(\alpha, \alpha)$, and use $b$ as the probability to generate a Bernoulli distribution, from which we sample $\lambda \in \mathbb{R}^{B \times C}$($\lambda \sim \operatorname{Bern}(P)$), and based on \cite{zhou2021domain}, $\alpha$ is empirically set to 0.1. The augmented statistics $\mu^{\prime},\sigma^{\prime}$ and original statistics $\mu,\sigma$ are then mixed along the channels:
\begin{equation}
	\begin{aligned}
		 \beta_{\operatorname{mix}}=\lambda \mu+(1-\lambda) \mu^{\prime} ,\gamma_{\operatorname{mix}}=\lambda \sigma+(1-\lambda) \sigma^{\prime},
	\end{aligned}
\end{equation}
where $\mu,\sigma$ are derived from the shallow features $f$ of the encoder. Finally, the mixed feature statistics are applied to perturb the normalized $f$:
\begin{equation}
	\begin{aligned}
		 \operatorname{AFB}(f)=\gamma_{\operatorname{mix}} \frac{f-\mu}{\sigma}+\beta_{\operatorname{mix}}.
	\end{aligned}
\end{equation}
Compared to MixStyle\cite{zhou2021domain}, we use augmentation statistics randomly sampled from a uniform distribution that encompasses most feature statistics, instead of randomly selecting augmentation statistics from the source domain. This approach generates out-of-distribution samples, significantly enhancing the diversity of the training data. Additionally, by incorporating references from the original features, which could prevent the model from overgeneralizing or failing to converge due to excessive randomization.

\subsection{Dual Cross-Attention Regularization}
After applying Adaptive Feature Blending, we obtain original deep features $f^{\prime}$ and generated deep features $f^{\prime}_{AFB}$ with different styles but the same semantics. To enable the model to learn more domain-invariant representations, we leverage the inherent long-term dependencies of the attention mechanism. We leverage the semantic information of the generated deep features and apply a cross-channel attention mechanism \cite{wang2022uctransnet,ding2022davit} to reconstruct the original deep features into robust regularization representations. Specifically, the original deep features are used as queries, while the generated deep features serve as keys and values. We calculate the similarity between each channel of the original and generated deep features, with the most similar channels of the generated deep features playing a more crucial role in reconstructing the original deep features. Given the hidden features  $\left[\bar{f}, \bar{f}_{AFB}\right]$ of the original and generated deep features after the encoding stage, we split them then set $\bar{f}$ as query and $\bar{f}{AFB}$ as key and value in multi-head manner:
\begin{equation}
    \label{lab3}
	\begin{aligned}
        q=\bar{f} w_{q}, k=\bar{f}_{AFB} w_{k}, v=\bar{f}_{AFB} w_{v},
	\end{aligned}
\end{equation}
where $w_{q}, w_{k}, w_{v} \in \mathbb{R}^{C \times 2C}$ are transformation weights, $\bar{f}, \bar{f}_{AFB} \in \mathbb{R}^{HW \times C}$, C is the channel dimension. The cross-channel attention is defined as:
\begin{equation}
    \label{lab4}
	\begin{aligned}
        \hat f=softmax\left[\psi\left(q^{\top} k\right)\right] v^{\top} w_{\text {out }}
	\end{aligned}
\end{equation}
where $\psi$ denote the instance normalization, $w_{\text {out }} \in \mathbb{R}^{2C \times C}$. While channel-wise attention can capture long-range dependencies between channels, we also apply channel self-attention to further refine the hidden features of the generated deep features. The formula for obtaining the refined generated features is similar to (\ref{lab3}) and (\ref{lab4}), with the only difference being that $q$ is replaced by $q=\bar{f}_{AFB} w_{q}$. 
This forms a display constraint by using the original deep features as queries to enforce consistency on the cross-domain generated deep feature representations, thereby encouraging the model to learn domain-invariant representations. Similarly, we can apply a display constraint by using the cross-domain generated deep features as queries to enforce consistency on the original deep feature representations. Finally, we fuse the reconstructed and self-attended original features as input to the decoder to generate the final prediction. The generated features undergo the same process.

\subsection{Loss Function}
We employ both the Dice loss \cite{milletari2016v} and the unified cross-entropy (CE) loss \cite{murphy2012machine}  for segmentation losses. To combat domain shift, we also use the mean squared error function as a semantic consistency loss term \cite{zhou2022generalizable}. The overall loss function is described as follows:
\begin{equation}
    \begin{aligned}
        \mathcal{L}_{{total}}=& \frac{1}{D} \sum_{d=1}^{D}(\mathcal{L}_{{seg}}^{d}+\mathcal{L}_{{consist}}^{d}).
    \end{aligned}
\end{equation}

\begin{table*}[!h]
\begingroup  
    \setlength{\tabcolsep}{1mm}  
	\centering
	\caption{ Dice coefficient (\%) and Average Surface Distance (voxel) produced by different methods on the Fundus dataset. The best results are bold-faced.}
	\label{tab1}
    \small
		\begin{tabular}{@{}c|cccc|c|cccc|c@{}}
		\toprule
		\multirow{2}{*}{Task} & \multicolumn{5}{c|}{Optic Cup/Disc Segmentation} & \multicolumn{5}{c@{}}{Optic Cup/Disc Segmentation} \\ \cmidrule(lr){2-6} \cmidrule(lr){7-11}
		& Domain 1 & Domain 2 & Domain 3 & Domain 4 & Avg & Domain 1 & Domain 2 & Domain 3 & Domain 4 & Avg \\ \midrule
		DoFE  & 79.66/95.95 & 79.62/88.42 & 85.45/90.53 & 83.68/\textbf{94.84} & 87.27 & 20.85/7.04 & 14.44/18.20 & 10.22/13.20 & 9.11/\textbf{5.81} & 12.36 \\
		HCDG  & 78.26/92.69 & 80.14/90.07 & 84.88/92.88 & 86.27/94.72 & 87.49 & 21.94/12.31 & 14.55/14.65 & 10.42/10.05 & 7.87/5.85 & 12.21 \\
		TVConv  & 85.87/96.28 & 78.17/91.72 & 86.86/92.19 & 82.04/91.95 & 88.14 & 14.78/6.51 & 15.60/14.84 & 9.33/10.95 & 10.17/10.13 & 11.54 \\
		RAM-DSIR  & 83.08/95.59 & 76.98/87.30 & 84.91/94.86 & 84.45/93.65 & 87.60 & 17.92/7.90 & 16.99/18.84 & 10.51/7.38 & 9.18/7.49 & 12.03 \\
		CDDSA & 85.78/\textbf{96.36} & 78.38/91.81 & 86.26/\textbf{95.10} & 83.18/92.59 & 88.68 & 14.84/\textbf{6.33} & 14.48/11.84 & 9.73/\textbf{7.04} & 9.68/8.13 & 10.26 \\
		WT-PSE & \textbf{86.71}/96.23 & 79.64/\textbf{92.67} & \textbf{87.90}/93.69 & 83.75/93.04 & 89.20 & 14.10/6.76 & \textbf{13.46}/10.50 & 8.97/8.84 & 9.49/8.85 & 10.12 \\ \midrule
		Ours & 85.86/95.89 & \textbf{80.76}/89.42 & 86.46/94.96 & \textbf{87.62}/94.58 & \textbf{89.44} & \textbf{13.29}/8.15 & 13.58/\textbf{8.93} & \textbf{8.08}/8.29 & \textbf{7.19}/5.96 & \textbf{9.18} \\
		\bottomrule
	\end{tabular}
    \normalsize
\endgroup  
\end{table*}

\begin{table*}[!h]
	\centering
	\caption{\centering Dice coefficient (\%) and Average Surface Distance (voxel) produced by different methods on the Prostate dataset. The best results are bold-faced.}
	\label{tab2}
	\small
		\begin{tabular}{@{}c|cccccc|c@{}}
		\toprule
		Task     & \multicolumn{6}{c|}{Prostate Segmentation(Dice coefficient / Average Surface Distance)}   & \multirow{2}{*}{\centering Avg} \\ \cmidrule(r){1-7}
		Domain   & Domain 1 & Domain 2 & Domain 3 & Domain 4 & Domain 5 & Domain 6 & \multicolumn{1}{c}{}                     \\ \midrule
		DoFE   & 85.23/2.31    & 88.16/1.56    & 85.98/2.51    & 87.34/2.86    & 85.64/2.69    & 82.39/3.04    & 85.79/2.50                                    \\
		HCDG     & 85.31/2.28    & 88.43/1.49    & 84.54/2.74    & 88.37/1.88    & 84.12/2.76    & 86.36/2.20    & 86.19/2.23                                    \\
		TVConv  & 86.59/2.25    & 88.75/1.65    & 85.89/2.36    & 88.63/1.74    & 85.38/2.17    & 87.94/1.41    & 87.20/1.93                                    \\
		RAM-DSIR & 87.19/2.07    & 88.92/1.59    & 86.13/\textbf{2.04}    & 88.46/1.56    & 87.26/1.82    & 87.67/1.33    & 87.61/1.74                                    \\
		CDDSA  & 88.65/1.84    & 88.91/1.76    & 86.58/2.13    & 88.35/1.48    & 86.71/1.98    & 88.24/1.16    & 87.91/1.73                                    \\
		WT-PSE & 89.65/1.23    & 89.34/1.19    & 86.57/2.21    & 89.74/1.01    & \textbf{88.25}/\textbf{1.41}    & 89.97/1.08    & 88.92/1.41                                   \\ \midrule
		Ours     & \textbf{90.66}/\textbf{0.84}    & \textbf{90.73}/\textbf{0.73}    & \textbf{87.83}/2.74    & \textbf{90.02}/\textbf{0.61}    & 88.16/1.58    & \textbf{90.26}/\textbf{0.77}    & \textbf{89.61}/\textbf{1.21}                                    \\
		\bottomrule
	\end{tabular}
	\normalsize
\end{table*}

\section{Experiment}
\subsection{Datasets and Implementation Details }
Our experiments use two publicly available medical image segmentation datasets: the Fundus \cite{wang2020dofe} and Prostate \cite{liu2020shape} datasets. The Fundus dataset includes retinal fundus images from four different medical centers, used for optic cup and disc segmentation. The Prostate dataset consists of T2-weighted MRI images from six sources, focused on prostate segmentation. The data preprocessing strictly follows previous studies \cite{zhou2022generalizable}.

We utilize a U-shaped segmentation network \cite{hu2022domain} built on the ResNet-34 \cite{he2016deep} backbone and conduct our experiments on an NVIDIA 3090 GPU. The batch size is set to 8, and we train for 200 epochs using an SGD optimizer with a momentum of 0.99 for both tasks. The initial learning rates are set to 0.001 for the prostate dataset and 0.0007 for the OD/OC dataset. To stabilize the training process, the learning rate decays according to a polynomial schedule. For both tasks, we train on $K-1$ source domains and evaluate on the remaining target domain (with a total of $K$ domains). The evaluation metrics include the Dice coefficient (Dice) and Average Surface Distance (ASD), following the standards set by previous medical image segmentation studies. To reduce randomness, we repeated the experiments three times and reported the average performance.

\subsection{Comparison against other methods}

\begin{figure}
	\centering
	\includegraphics[width=\columnwidth]{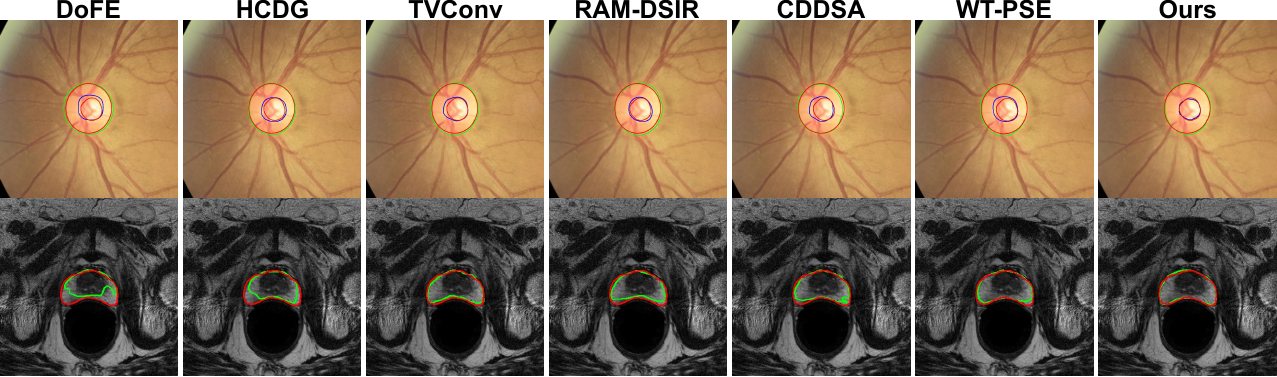}
	\caption{Visual comparison for Fundus and Prostate segmentation task. The red contours indicate the boundaries of ground truths while the green and blue contours are predictions.}
	\label{fig3}
\end{figure}

We compared our method with six state-of-the-art domain generalization methods: DoFE\cite{wang2020dofe}, HCDG\cite{yang2021hcdg}, TVConv\cite{chen2022tvconv}, RAM-DSIR\cite{zhou2022generalizable}, CDDSA\cite{gu2023cddsa}, and WT-PSE\cite{chen2024learning}. Tables \ref{tab1} and \ref{tab2} show the Dice coefficients and ASD results across different domains on the Fundus and Prostate datasets. Our method consistently achieves higher Dice coefficients and better ASD scores, demonstrating its effectiveness. Additionally, Fig. \ref{fig3} provides visualizations from the target domain, showing that our method outperforms others with clearer boundary segmentation.

\setlength{\tabcolsep}{0.5mm}
\begin{table}[t]
	\centering
	\caption{Ablation Study of key components in our method on the Fundus dataset (\%). }
	\label{tab3}
    \small
		\begin{tabular}{@{}cc|cccc|c@{}}
        \toprule
        \multicolumn{2}{c|}{Task} & \multicolumn{4}{c|}{Optic Cup/Disc Segmentation}      & \multirow{2}{*}{Avg} \\ \cmidrule(r){1-6}
        AFB         & DCAR        & Domain 1    & Domain 2    & Domain 3    & Domain 4    &                      \\ \midrule
        -           & -           & 70.19/93.07 & 71.11/88.59 & 82.46/91.98 & 81.63/92.14 & 83.90                \\
        \checkmark  & -           & 80.85/94.78 & 77.21/87.84 & 84.51/94.01 & 84.79/92.99 & 87.12                \\
        \checkmark  & \checkmark  & 85.86/95.89 & 80.76/89.42 & 86.46/94.96 & 87.62/94.58 & 89.44                \\ \bottomrule
        \end{tabular}
    \normalsize
\end{table}

\subsection{Ablation study}
We performed ablation studies to assess the effectiveness of different components in our method. As shown in Table \ref{tab3}, each component contributed to the overall performance improvement. In particular, Adaptive Feature Blending(AFB) provided a benefit of 3.22, while Dual Cross-Attention Regularization(DCAR) resulted in a benefit of 2.32. These findings demonstrate that AFB and DCAR successfully regularize our segmentation model and enhance its generalization capability.

\section{Conclusion}
This paper proposes a data augmentation method called Adaptive Feature Blending, which not only explores the in-distribution space but also generates out-of-distribution samples, significantly extending the domain's coverage. Additionally, we introduce a plug-and-play module named Dual Cross-Attention Regularization for domain-generalized medical image segmentation, aimed at guiding the model to learn domain-invariant representations. We innovatively use deep features combined with a cross-attention mechanism to form regularization constraints, steering the model towards learning domain-invariant representations. Extensive experiments demonstrate that the proposed method achieves state-of-the-art performance.

\bibliographystyle{IEEEtran}
\bibliography{conference_101719}
\vspace{12pt}
\color{red}

\end{document}